\documentclass{article}

\usepackage[T1]{fontenc}

\usepackage[utf8]{inputenc} \usepackage[T1]{fontenc}
\usepackage{tgtermes}
\usepackage{amssymb}
\usepackage{amsfonts}

\usepackage{tgtermes}
\usepackage{amsmath}

\usepackage{scalefnt,letltxmacro}
\LetLtxMacro{\oldtextsc}{\textsc}
\renewcommand{\textsc}[1]{\oldtextsc{\scalefont{1.10}#1}}
\usepackage[scaled=0.92]{PTSans}

\usepackage[usenames,dvipsnames]{xcolor}
\definecolor{shadecolor}{gray}{0.9}

\usepackage[parfill]{parskip}
\usepackage{afterpage}
\usepackage{framed}
\usepackage{nicefrac}

\newcommand{\mathbold}[1]{\ensuremath{\boldsymbol{\mathbf{#1}}}}

\usepackage[colorinlistoftodos,
           prependcaption,
           textsize=small,
           backgroundcolor=yellow,
           linecolor=lightgray,
           bordercolor=lightgray]{todonotes}

\usepackage{lineno}

\usepackage{ragged2e}

\DeclareRobustCommand{\parhead}[1]{\textbf{#1}~}

\usepackage{graphicx}
\usepackage[labelfont=bf]{caption}
\usepackage[format=hang]{subcaption}

\usepackage{booktabs}
\usepackage{arydshln} 
\usepackage{natbib}

\usepackage{listings}
\usepackage{fancyvrb}
\fvset{fontsize=\normalsize}

\usepackage[colorlinks,linktoc=all]{hyperref}
\usepackage[all]{hypcap}
\hypersetup{citecolor=Violet}
\hypersetup{linkcolor=black}
\hypersetup{urlcolor=MidnightBlue}
\usepackage{url}

\usepackage[acronym,smallcaps,nowarn]{glossaries}

\usepackage{listings}
\lstdefinestyle{alp_style}{
    commentstyle=\color{OliveGreen},
    numberstyle=\tiny\color{black!60},
    stringstyle=\color{BrickRed},
    basicstyle=\ttfamily\scriptsize,
    breakatwhitespace=false,
    breaklines=true,
    captionpos=b,
    keepspaces=true,
    numbers=none,
    numbersep=5pt,
    showspaces=false,
    showstringspaces=false,
    showtabs=false,
    tabsize=2
}
\newcommand{\g}{\,|\,}

\usepackage{bbold}

\usepackage{mathtools}

\newcommand{\mbc}{\mathbold{c}}

\newcommand{\mbh}{\mathbold{h}}

\newcommand{\mbs}{\mathbold{s}}

\newcommand{\mbx}{\mathbold{x}}

\newcommand{\mbz}{\mathbold{z}}

\newcommand{\mbI}{\mathbold{I}}

\newcommand{\mbepsilon}{\mathbold{\epsilon}}

 \newacronym{VI}{vi}{variational inference}
\newacronym{KL}{kl}{Kullback-Leibler}
\newacronym{ELBO}{elbo}{\emph{evidence lower bound}}
\newacronym{MCMC}{mcmc}{Markov chain Monte Carlo}
\newacronym{VAE}{vae}{\emph{variational autoencoder}}
\newacronym{RNN}{rnn}{recurrent neural network}
\newacronym{MLP}{mlp}{\emph{feed forward neural network}}
\newacronym{DEF}{def}{\emph{deep exponential family}}
\newacronym{LSTM}{lstm}{Long-Short Term Memory}
\newacronym{GRU}{gru}{Gated Recurrent Unit}
\newacronym{VRNN}{vrnn}{\emph{Variational Recurrent Neural Network}}
\newacronym{SRNN}{srnn}{\emph{Stochastic Recurrent Neural Network}}
\newacronym{ELMAN}{elman}{\emph{Elman}}
\newacronym{ERNN}{ernn}{Elman Recurrent Neural Network}
\newglossaryentry{noisin}
{name=noisin,
  description={unbiased regularization with noise injection}
}
 
\usepackage[accepted]{icml2018}

\hypersetup{citecolor=Violet}
\hypersetup{linkcolor=black}
\hypersetup{urlcolor=MidnightBlue}

\icmltitlerunning{\Gls{noisin}: Unbiased Regularization for Recurrent Neural Networks}

\begin{document}

\twocolumn[
\icmltitle{\Gls{noisin}: Unbiased Regularization for Recurrent Neural Networks}

\icmlsetsymbol{equal}{*}

\begin{icmlauthorlist}
\icmlauthor{Adji B. Dieng}{columbia}
\icmlauthor{Rajesh Ranganath}{nyu}
\icmlauthor{Jaan Altosaar}{princeton}
\icmlauthor{David M. Blei}{columbia}
\end{icmlauthorlist}

\icmlaffiliation{columbia}{Columbia University}
\icmlaffiliation{nyu}{New York University}
\icmlaffiliation{princeton}{Princeton University}

\icmlcorrespondingauthor{Adji B. Dieng}{abd2141@columbia.edu}

\icmlkeywords{regularization, recurrent neural networks, language modeling, learning theory, deep learning}

\vskip 0.3in
]

\printAffiliationsAndNotice{}  

\begin{abstract}
  \vskip 0.1in \Glspl{RNN} are powerful models of sequential
  data.~They have been successfully used in domains such as text and
  speech.~However, \gls{RNN}s are susceptible to overfitting;
  regularization is important. In this paper we develop \Gls{noisin},
  a new method for regularizing \gls{RNN}s.  \Gls{noisin} injects
  random noise into the hidden states of the \gls{RNN} and then
  maximizes the corresponding marginal likelihood of the data. We show
  how \Gls{noisin} applies to any \gls{RNN} and we study many
  different types of noise. \Gls{noisin} is unbiased---it preserves
  the underlying \gls{RNN} on average.  We characterize how
  \Gls{noisin} regularizes its \gls{RNN} both theoretically and
  empirically. On language modeling benchmarks, \Gls{noisin} improves
  over dropout by as much as $12.2\%$ on the Penn Treebank and
  $9.4 \%$ on the Wikitext-2 dataset.  We also compared the
  state-of-the-art language model of Yang et al. 2017, both with and
  without \Gls{noisin}.  On the Penn Treebank, the method with Noisin
  more quickly reaches state-of-the-art performance.
\end{abstract}

\section{Introduction}\label{sec:introduction}
\glsresetall

\Glspl{RNN} are powerful models of sequential data
\citep{robinson1987utility, werbos1988generalization,
  williams1989complexity, elman1990finding,
  pearlmutter1995gradient}.~\gls{RNN}s have achieved state-of-the-art 
  results on many tasks, including language modeling \citep{mikolov2012context,
  yang2017breaking}, text generation \citep{graves2013generating},
image generation \citep{gregor2015draw}, speech recognition
\citep{graves2013speech, chiu2017state}, and machine translation
\citep{sutskever2014sequence,wu2016google}.

The main idea behind an \gls{RNN} is to posit a sequence of
recursively defined \emph{hidden states}, and then to model each
observation conditional on its state.  The key element of an \gls{RNN}
is its \textit{transition function}.  The transition function
determines how each hidden state is a function of the previous
observation and previous hidden state; it defines the underlying
recursion. There are many flavors of \gls{RNN}s---examples include the
\gls{ERNN} \citep{elman1990finding}, the \gls{LSTM}
\citep{hochreiter1997long}, and the \gls{GRU} \citep{cho2014learning}.
Each flavor amounts to a different way of designing and parameterizing
the transition function.

We fit an \gls{RNN} by maximizing the likelihood of the observations
with respect to its parameters, those of the transition function and
of the observation likelihood. But \gls{RNN}s are very flexible and
they overfit; regularization is crucial. Researchers have explored
many approaches to regularizing \gls{RNN}s, such as Tikhonov
regularization \citep{bishop1995training}, dropout and its variants
\citep{srivastava2014dropout, zaremba2014recurrent,
  gal2016theoretically, wan2013regularization}, and zoneout
\citep{krueger2016zoneout}.  (See the related work section below for
more discussion.)

In this paper, we develop \Gls{noisin}, an effective new way to
regularize an \gls{RNN}. The idea is to inject random noise into its
transition function and then to fit its parameters to maximize the
corresponding marginal likelihood of the observations. We can easily
apply \Gls{noisin} to any flavor of \gls{RNN} and we can use many
types of noise.

\begin{figure}[t]
\centering
 	\includegraphics[width=\columnwidth, height=6.0cm]{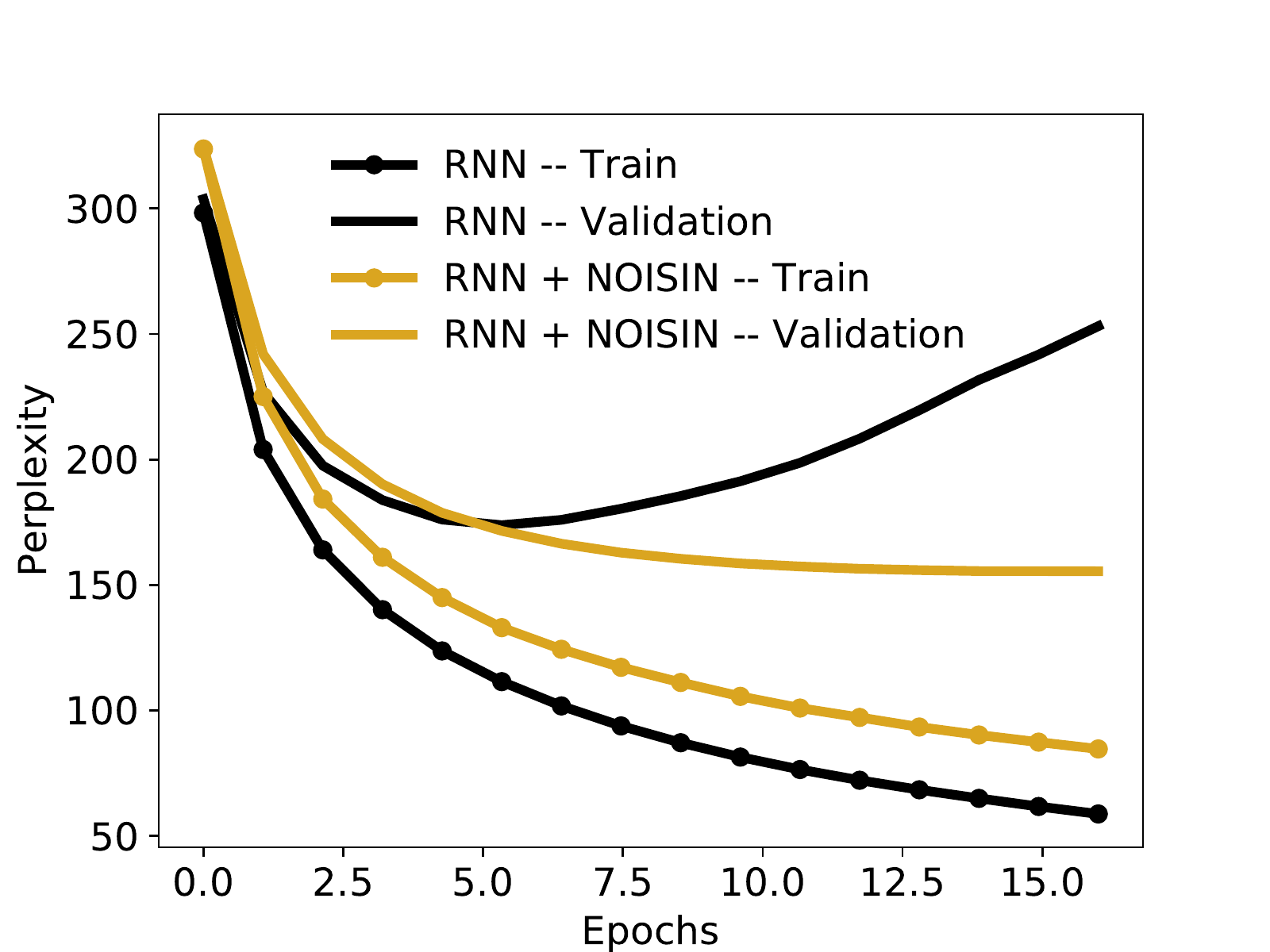}
	\caption{Training and validation perplexity for the deterministic \gls{RNN} and
	the \gls{RNN} regularized with \Gls{noisin}.~The settings were the same for both. 
	We used additive Gaussian noise on an \gls{ERNN} with sigmoid activations. 
	We used one layer of $256$ hidden units. The \gls{RNN} overfits after only five 
	epochs, and its training loss still decreases. This is not the case for the \gls{RNN} 
	regularized with \Gls{noisin}. }
	\label{fig:pp_plot}
\end{figure}

Figure\nobreakspace \ref {fig:pp_plot} demonstrates how an \gls{RNN} can overfit and how
\Gls{noisin} can help. The plot involves a language modeling task
where the \gls{RNN} models a sequence of words.  The horizontal axis
is epochs of training; the vertical axis is perplexity, which is an
assessment of model fitness (lower numbers are better).~The figure
shows how the model fits to both the training set and the validation
set.~As training proceeds, the vanilla \gls{RNN} improves its fitness
to the training set but performance on the validation set
degrades---it overfits. The performance of the \gls{RNN} with \Gls{noisin} 
continues to improve in both the training set and the validation set.

\Gls{noisin} regularizes the \gls{RNN} by smoothing its loss,
averaging over local neighborhoods of the transition function.
Further, \Gls{noisin} requires that the noise-injected transition
function be \textit{unbiased}.  This means that, on average, it
preserves the transition function of the original \gls{RNN}.

With this requirement, we show that \Gls{noisin} provides
explicit regularization, i.e., it is equivalent to fitting the usual
\gls{RNN} loss plus a penalty function of its parameters. We can
characterize the penalty as a function of the variance of the noise.
Intuitively, it penalizes the components of the model that are
sensitive to noise; this induces robustness to how future data may be
different from the observations.

We examine \Gls{noisin} with the \gls{LSTM} and the 
\gls{LSTM} with dropout, which we call the dropout-\gls{LSTM}, and we
explore several types of distributions. We study performance with two 
benchmark datasets on a language modeling task. \Gls{noisin} improves 
over the \gls{LSTM} by as much as $37.3\%$ on the
Penn Treebank dataset and $39.0\%$ on the Wikitext-2 dataset; it improves 
over the dropout-\gls{LSTM} by as much as $12.2\%$ on the Penn Treebank 
and $9.4\%$ on Wikitext-2.

\parhead{Related work.}~Many techniques have been developed to address
overfitting in \gls{RNN}s.~The most traditional regularization
technique is weight decay ($L_1$ and $L_2$). However,
\citet{pascanu2013difficulty} showed that such simple regularizers
prevent the \glspl{RNN} from learning long-range dependencies.

Another technique for regularizing \gls{RNN}s is to normalize the
hidden states or the observations~\citep{ioffe2015batch, ba2016layer,
  cooijmans2016recurrent}.  Though powerful, this class of approaches
can be expensive.

Other types of regularization, including what we study in this paper,
involve auxiliary noise variables. The most successful noise-based
regularizer for neural networks is
dropout~\citep{srivastava2014dropout, wager2013dropout,
  noh2017regularizing}. Dropout has been adapted to \gls{RNN}s by only
pruning their input and output matrices \citep{zaremba2014recurrent}
or by putting judiciously chosen priors on all the weights and
applying variational methods \citep{gal2016theoretically}.  Still
other noise-based regularization prunes the network by dropping
updates to the hidden units of the \gls{RNN}~
\citep{krueger2016zoneout, semeniuta2016recurrent}. More recently
\citet{merity2017regularizing} extended these techniques.

Involving noise variables in \gls{RNN}s has been used in contexts
other than regularization.  For example \citet{jim1996analysis}
analyze the impact of noise on convergence and long-term dependencies.
Other work introduces auxiliary latent variables that enable
\gls{RNN}s to capture the high variability of complex sequential data
such as music, audio, and text~\citep{bayer2014learning,
  chung2015recurrent, fraccaro2016sequential, goyal2017z}. \section{Recurrent Neural Networks}\label{sec:background}

\begin{table*}[t]\label{tab:A}
\centering
\caption{Expression for the log normalizer $A$ and its Hessian $\nabla^2 A$ for different likelihoods. 
 Here $\sigma^2$ is the observation variance in the Gaussian case and $\eta = \exp(\mbs)$ in the categorical case.}
\begin{tabular}{l|c|c}
\toprule
Likelihood & $A(\mbs)$ & $\nabla^2 A(\mbs)$ \\\midrule
Bernoulli (Binary data) & $-\log(1 - \sigma(\mbs))$ & $ \sigma(\mbs) \cdot (1 - \sigma(\mbs))$ \\
Gaussian (Real-Valued data)  & $\frac{1}{2 \sigma^2} \mbs^\top \mbs$ & $\frac{1}{\sigma^2} \mbI$ \\
Poisson (Count data)  &$\exp(\mbs)$ & $\exp(\mbs)$\\
Categorical (Categorical data) & logsumexp$(\eta)$ & $\frac{1}{\mathbb{1}^\top \eta} \text{diag}(\eta) - \frac{\eta \eta^\top}{(\mathbb{1}^\top \eta)^2}$ 
\\\bottomrule
\end{tabular}\label{tab:A}
\end{table*}

Consider a sequence of observations, $\mbx_{1:T} = (\mbx_1, ... ,\mbx_T)$. An \gls{RNN}
factorizes its joint distribution according to the chain rule of probability,
\begin{align}
  \label{eq:chain_rule}
  p(\mbx_{1:T}) = \prod_{t=1}^{T} p(\mbx_t | \mbx_{1: t-1}).
\end{align}
To capture dependencies, the \gls{RNN} expresses each conditional
probability as a function of a low-dimensional recurrent hidden state,
\begin{align*}  \mbh_t  &= f_W(\mbx_{t-1}, \mbh_{t-1}) \text{ and }
  p(\mbx_{t} | \mbx_{1: t-1}) = p(\mbx_{t} | \mbh_t).
\end{align*}

The likelihood $p(\mbx_{t} | \mbh_t)$ can be 
of any form. We focus on the exponential family 
\begin{align}\label{eq:likelihood}
	p(\mbx_{t} | \mbh_t)  &= \nu(\mbx_t) \exp \left\{(V^\top \mbh_t)^\top \mbx_t - A(V^\top \mbh_t) \right\}
	,
\end{align}
where $\nu(\cdot)$ is the base measure, $V^\top \mbh_t$ is the natural 
parameter---a linear function of the hidden state $\mbh_t$---and $A(V^\top \mbh_t)$ is the 
log-normalizer. The matrix $V$ is called the \emph{prediction} or \emph{output} matrix 
of the \gls{RNN}. 

The hidden state $\mbh_t$ at time $t$ is a parametric
  function $f_W(\mbh_{t-1}, \mbx_{t-1})$ of the previous hidden state
$\mbh_{t - 1}$ and the previous observation $\mbx_{t - 1}$; the
parameters $W$ are shared across all time steps. The function $f_W$ is 
the transition function of the \gls{RNN}, it defines a recurrence relation 
for the hidden states and renders $\mbh_t$ a function of all the past 
observations $\mbx_{1:t-1}$; these properties match the chain rule 
decomposition in Eq.\nobreakspace \ref {eq:chain_rule}. 

The particular form of $f_W$ determines the \gls{RNN}. Researchers
have designed many flavors, including the \gls{LSTM} and 
the \gls{GRU} \citep{hochreiter1997long, cho2014learning}. In this 
paper we will study the \gls{LSTM}.~However, the 
methods we develop can be applied to all types of \gls{RNN}s.

\parhead{\acrlong{LSTM}.} 
We now describe the \gls{LSTM}, a variant of \gls{RNN} that we study 
in Section\nobreakspace \ref {sec:empirical}. The \gls{LSTM} is built from the simpler
\gls{ERNN}~\citep{elman1990finding}.~In an \gls{ERNN}, the transition
function is
\begin{align*}
  f_W(\mbx_{t-1}, \mbh_{t-1}) =
  s(W_x^\top \mbx_{t-1} + W_h^\top \mbh_{t-1})
  ,
\end{align*}
where we dropped an intercept term to avoid cluttered notation. Here,
$W_h$ is called the \emph{recurrent weight matrix} and $W_x$ is
called the \emph{embedding matrix} or \emph{input matrix}.~The function $s(\cdot)$ is 
called an \emph{activation} or \emph{squashing} function, which stabilizes the
transition dynamics by bounding the hidden state. Typical choices for
the squashing function include the sigmoid and the hyperbolic tangent.

The \gls{LSTM} was designed to avoid optimization issues, such as
vanishing (or exploding) gradients.  Its transition function composes
four \gls{ERNN}s, three with sigmoid activations and one with a
$\tanh$ activation:
\begin{gather}
  \label{eq:lstm-f}
  f_t = \sigma(W_{x1}^\top\mbx_{t-1} + W_{h1}^\top\mbh_{t-1}) \\
  i_t = \sigma(W_{x2}^\top\mbx_{t-1} + W_{h2}^\top\mbh_{t-1})\\
  o_t = \sigma(W_{x4}^\top\mbx_{t-1} + W_{h4}^\top\mbh_{t-1}) \\
  \mbc_t = f_t \odot \mbc_{t-1} + i_t \odot \tanh(W_{x3}^\top\mbx_{t-1} + W_{h3}^\top\mbh_{t-1}) \\
  \mbh_t = o_t \odot \tanh(\mbc_t) \label{eq:lstm_state}.
\end{gather}
The idea is that the memory cell $\mbc_t$ captures long-term
dependencies \citep{hochreiter1997long}. 

However, \gls{LSTM}s have a high model complexity and, consequently,
they easily memorize data. Regularization is crucial. In the next section, we develop a new regularization 
method for \glspl{RNN} called \Gls{noisin}. \section{Noise-Injected \gls{RNN}s}\label{sec:nirnns}
\glsresetall

\Gls{noisin} is built from noise-injected \gls{RNN}s. These are \gls{RNN}s 
whose hidden states are computed using auxiliary noise variables.~There 
are several advantages to injecting noise into the hidden states of \glspl{RNN}. 
For example it prevents the dimensions of the hidden states from co-adapting 
and forces individual units to capture useful features. 

We define noise-injected \gls{RNN}s as any \gls{RNN} following the 
generative process 
\begin{align}\label{eq:noisy-transition}
  \mbepsilon_{1:T} &\sim \varphi(\cdot; \mu, \gamma) \\
  \mbz_t &= g_W(\mbx_{t-1}, \mbz_{t-1}, \mbepsilon_t) \\
  p(\mbx_t \g \mbx_{1:t-1}) &= p(\mbx_t \g \mbz_t)
  ,
\end{align}
where the likelihood $p(\mbx_t \g \mbz_t)$ is an exponential family as in Eq.\nobreakspace \ref {eq:likelihood}. 
The noise variables $\mbepsilon_{1:T}$ are drawn from a distribution $\varphi(\cdot; \mu, \gamma)$ 
with mean $\mu$ and scale $\gamma$. For example, $\varphi(\cdot; \mu, \gamma)$ can be a zero-mean 
Gaussian with variance $\gamma^2$.~We will study many types of noise distributions.

The noisy hidden state $\mbz_t$ is a parametric function $g_W$ of the previous 
observation $\mbx_{t-1}$, the previous noisy hidden state $\mbz_{t-1}$, and the 
noise $\mbepsilon_t$. Therefore conditional on the noise $\mbepsilon_{1:T}$, the transition 
function $g_W$ defines a recurrence relation on $\mbz_{1:T}$. 

The function $g_W$ determines the noise-injected \gls{RNN}.~In this paper, we 
propose functions $g_W$ that meet the criterion described below. 

\parhead{Unbiased noise injection.}
Injecting noise at each time step limits the amount of information carried by 
hidden states. In limiting their capacity, noise injection is some form of 
regularization.~In Section\nobreakspace \ref {sec:reg} we show that noise injection under exponential 
family likelihoods corresponds to explicit regularization under some 
\textit{unbiasedness} condition. 

We define two flavors of unbiasedness: \textit{strong unbiasedness} and 
\textit{weak unbiasedness}. Let $\mbz_{t}(\mbepsilon_{1:t})$
denote the unrolled recurrence at time $t$; it is a random variable
via the noise $\mbepsilon_{1:t}$. Under the strong unbiasedness condition, the transition 
function $g_W$ must satisfy the relationship 
\begin{align}\label{eq:unbiasedness}
  \mathbb{E}_{p(\mbz_t(\mbepsilon_{1:t}) \g \mbz_{t-1})}\left[\mbz_t(\mbepsilon_{1:t})\right]
  &= \mbh_t
\end{align}
where $\mbh_t$ is the hidden state of the underlying \gls{RNN}. 
This is satisfied by injecting the noise at the last layer of the \gls{RNN}. 
Weak unbiasedness imposes a looser constraint. Under weak unbiasedness, $g_W$ must satisfy 
 \begin{align}\label{eq:unbiasedness}
  \mathbb{E}_{p(\mbz_t(\mbepsilon_{1:t}) \g \mbz_{t-1})}\left[\mbz_t(\mbepsilon_{1:t})\right]
  &= f_W(\mbx_{t-1}, \mbz_{t-1})
\end{align}
where $f_W$ is the transition function of the underlying \gls{RNN}. 
What weak unbiasedness means is that the noise should be injected in such a way 
that driving the noise to zero leads to the original \gls{RNN}.~Two possible choices 
for $g_W$ that meet this condition are the following
\begin{align}
  g_W(\mbx_{t-1}, \mbz_{t-1}, \mbepsilon_t) &= f_W(\mbx_{t-1}, \mbz_{t-1}) + \mbepsilon_t \label{eq:additive_noise} \\
  g_W(\mbx_{t-1}, \mbz_{t-1}, \mbepsilon_t) &= f_W(\mbx_{t-1}, \mbz_{t-1}) \odot \mbepsilon_t.\label{eq:multiplicative_noise}
\end{align}
In Eq.\nobreakspace \ref {eq:additive_noise} the noise has mean zero whereas in Eq.\nobreakspace \ref {eq:multiplicative_noise} it has mean 
one.~These choices of $g_W$ correspond to additive noise and multiplicative noise respectively. 
Note $f_W$ can be any \gls{RNN} including the \gls{RNN} with dropout or the 
stochastic \gls{RNN}s \citep{bayer2014learning, chung2015recurrent, fraccaro2016sequential, goyal2017z}.
For example to implement unbiased noise injection with multiplicative noise for 
the \gls{LSTM} the only change from the original \gls{LSTM} is to replace Eq.\nobreakspace \ref {eq:lstm_state} with 
\begin{align*}
  \mbz_t = o_t \odot \tanh(\mbc_t) \odot \mbepsilon_t.
\end{align*}
Such noise-injected hidden states can be stacked to build a multi-layered noise-injected \gls{LSTM} that 
meet the weak unbiasedness condition. 

\parhead{Dropout.} We now consider dropout from the perspective of unbiasedness. 
Consider the \gls{LSTM} as described in Section\nobreakspace \ref {sec:background}.~Applying dropout to it corresponds to 
injecting Bernoulli-distributed noise as follows 
\begin{gather*}
  \label{eq:lstm-f}
  f_t = \sigma(W_{x1}^\top\mbx_{t-1} \odot \mbepsilon^{xf}_t + W_{h1}^\top\mbh_{t-1} \odot \mbepsilon^{hf}_t) \\
  i_t = \sigma(W_{x2}^\top\mbx_{t-1}\odot \mbepsilon^{xi}_t + W_{h2}^\top\mbh_{t-1} \odot \mbepsilon^{hi}_t)\\
  o_t = \sigma(W_{x4}^\top\mbx_{t-1} \odot \mbepsilon^{xo}_t+ W_{h4}^\top\mbh_{t-1} \odot \mbepsilon^{ho}_t) \\
  \mbc_t = f_t \odot \mbc_{t-1} +\\
  		 i_t \odot \tanh(W_{x3}^\top\mbx_{t-1} \odot \mbepsilon^{xc}_t 
		 + W_{h3}^\top\mbh_{t-1} \odot \mbepsilon^{hc}_t) \\
  \mbz^{dropout}_t = o_t \odot \tanh(\mbc_t) .
\end{gather*}
This general form of dropout encapsulates existing dropout variants. For example when 
the noise variables $\mbepsilon^{hf}_t, \mbepsilon^{hi}_t, \mbepsilon^{ho}_t, \mbepsilon^{hc}_t$ are 
set to one we recover the variant of dropout in \citet{zaremba2014recurrent}. 

Because of the nonlinearities dropout does not meet the unbiasedness desideratum Eq.\nobreakspace \ref {eq:unbiasedness}
where $\mbh_t$ is the hidden state of the \gls{LSTM} as described in Section\nobreakspace \ref {sec:background}. 
Here at each time step $t$, $\mbepsilon_t$ denotes the set of noise variables 
$\mbepsilon^{xf}_t, \mbepsilon^{xi}_t, \mbepsilon^{xo}_t, \mbepsilon^{xc}_t$ 
and $\mbepsilon^{hf}_t, \mbepsilon^{hi}_t, \mbepsilon^{ho}_t, \mbepsilon^{hc}_t$. 

Dropout is therefore biased and does not preserve the underlying \gls{RNN}. 
However, dropout has been widely successfully used in practice and has many nice properties. 
For example it regularizes by acting like an ensemble method~\citep{goodfellow2016deep}. We 
study the dropout-\gls{LSTM} in Section\nobreakspace \ref {sec:empirical} as 
a variant of \gls{RNN} that can benefit from the method \Gls{noisin} proposed in this paper.

\newlength{\textfloatsepsave}
\setlength{\textfloatsepsave}{\textfloatsep}
\setlength{\textfloatsep}{0.10in} 

\begin{algorithm}[tb]
	\caption{\Gls{noisin} with multiplicative noise.}
	\label{alg:noisin}
	\begin{algorithmic}
		\STATE {\bfseries Input:} Data $\mbx_{1:T}$, initial hidden state $\mbz_0$, noise 
 distribution $\varphi(\cdot; 1, \gamma)$, and learning rate $\rho$.
		\STATE {\bfseries Output:} learned parameters $W^*$ and $V^*$.
		\STATE Initialize parameters $W$ and $V$
		\FOR{iteration $\text{iter}=1,2,\ldots,$}
			\FOR{time step $t=1, \ldots, T$}
				\STATE Sample noise $\mbepsilon_t \sim \varphi(\mbepsilon_t; 1, \gamma)$
				\STATE Compute state $\mbz_t = f_W(\mbz_{t-1}, \mbx_{t-1}) \odot \mbepsilon_t$
			\ENDFOR
			\STATE Compute loss $\widetilde{\mathcal{L}}$ as in Eq.\nobreakspace \ref {eq:final_objective}
			\STATE Update $W:\leftarrow W - \rho \cdot \nabla_{W} \widetilde{\mathcal{L}}$
			\STATE Update $V:\leftarrow V - \rho \cdot \nabla_{V} \widetilde{\mathcal{L}}$
			\STATE Change learning rate $\rho$ according to some schedule.
		\ENDFOR
	\end{algorithmic}
\end{algorithm}

\setlength{\textfloatsep}{\textfloatsepsave} 
\parhead{Unbiased noise-injection with \Gls{noisin}.}
Deterministic \gls{RNN}s are learned using truncated backpropagation through time 
with the maximum likelihood objective---the log likelihood of the data. 
Backpropagation through time builds gradients by \emph{unrolling} the \gls{RNN}
into a feed-forward neural network and applies backpropagation \citep{rumelhart1988learning}. The 
\gls{RNN} is then optimized using gradient descent or 
stochastic gradient descent~\citep{robbins1951stochastic}. 

\Gls{noisin} applies the same procedure to the expected log-likelihood 
under the injected noise, 
\begin{align}\label{eq:initial_objective}
	\mathcal{L}
		&=E_{p(\mbepsilon_{1:T})}\left[ \log p(\mbx_{1:T} | \mbz_{1:T}(\mbepsilon_{1:T}))\right] 
		.
\end{align}
In more detail this is 
\begin{align}\label{eq:objective}
	\mathcal{L}
		 &= \sum_{t=1}^{T}E_{p(\mbepsilon_{1:t})} \Big[ 
			 \log p(\mbx_t | \mbz_t(\mbepsilon_{1:t}))\Big] 
\end{align}
Notice this objective is a Jensen bound on the marginal log-likelihood of the data, 
\begin{align*}
	\mathcal{L} 
	\leq \log E_{p(\mbepsilon_{1:T})} \left[ 
		p(\mbx_{1:T} | \mbz_{1:T}(\mbepsilon_{1:T}))
	\right]  = \log p(\mbx_{1:T})
	.
\end{align*}
The expectations in the objective of Eq.\nobreakspace \ref {eq:objective} are intractable due to the nonlinearities 
in the model and the form of the noise distribution.~We approximate the objective using Monte Carlo;
\begin{align*}
	\widehat{\mathcal{L}}
		 &= \frac{1}{K}\sum_{k=1}^{K}\sum_{t=1}^{T} \left[ 
			 \log p(\mbx_t | \mbz_t(\mbepsilon^{(k)}_{1:t}))\right] 
	.
\end{align*}
When using one sample ($K=1$), the training procedure 
is just as easy as for the underlying \gls{RNN}.~The loss in this case, under the 
exponential family likelihood, becomes 
\begin{align}\label{eq:final_objective}
	\widetilde{\mathcal{L}}
		 &= -\sum_{t=1}^{T} \left[ 
			(V^\top \mbz_t(\mbepsilon_{1:t}))^\top \mbx_t - A(V^\top \mbz_t(\mbepsilon_{1:t}))
	\right] + c
	,
\end{align}
where $c = -\sum_{t=1}^{T} \log \nu(\mbx_t)$ is a constant 
that does not depend on the parameters. 
Algorithm $1$ summarizes the procedure for multiplicative noise.~The only 
change from traditional \gls{RNN} training is when updating the hidden state in lines $4$ and $5$.

\begin{table*}[t]
\centering
\caption{Expression for the noise distributions and their scaled version used in this paper. 
 Here $\gamma$ is the noise spread. It determines the amount of regularization. 
 For example it is the standard deviation for Gaussian noise 
 and the scale parameter for Gamma noise. The constant $\delta = 0.5772$ is the Euler-Mascheroni constant}. 
\begin{tabular}{l|ccccc}
\toprule
Standard Noise $\eta$ & $E(\eta)$ & $Var(\eta)$ & Scaled Noise $\mbepsilon$ & $E(\mbepsilon)$ & $Var(\mbepsilon)$  \\\midrule
$\mathcal{N}(0, \gamma)$ & 0 & $\gamma^2$ & $\eta$ & $0$ & $\gamma^2$ \\
Bernoulli$(\gamma)$ & $\gamma$ & $\gamma (1 - \gamma)$ & $\frac{\eta}{\gamma}$ & $1$ & $\frac{1 - \gamma}{\gamma}$\\
Gamma$(\alpha, \gamma)$ & $\alpha \gamma$ &  $\alpha \gamma^2$ & $\frac{\eta - \alpha \gamma}{\sqrt{\alpha}}$ & $0$ & $\gamma^2$ \\
Gumbel$(0, \gamma)$ & $\delta \gamma$ & $\frac{\pi^2 \gamma^2}{6}$ & $\frac{\sqrt{6} (\eta - \delta \gamma)}{\pi}$ & $0$ & $\gamma^2$\\
Laplace$(0, \gamma)$  & $0$ & $2 \gamma^2$ & $\frac{\eta}{\sqrt{2}}$ & $0$ & $\gamma^2$ \\
Logistic$(0, \gamma)$ & $0$ & $\frac{\pi^2 \gamma^2}{3}$ & $\frac{\sqrt{3} \eta}{\pi}$ & $0$ & $\gamma^2$ \\
Beta$(\alpha, \gamma)$ &  $\frac{\alpha}{\alpha + \gamma}$ & $\frac{\alpha \gamma}{(\alpha + \gamma)^2 (\alpha + \gamma + 1)}$  & $(\alpha + \gamma) \sqrt{\frac{\alpha + \gamma + 1}{\alpha}}(\eta - \frac{\alpha}{\alpha + \gamma})$ & $0$ & $\gamma$\\
Chi-Square$(\gamma)$ & $\gamma$ & $2 \gamma$ & $\frac{\eta - \gamma}{\sqrt{2}}$ & $0$ & $\gamma$
\\\bottomrule
\end{tabular}\label{tab:scaled_noise}
\end{table*}
 
\parhead{Controling the noise level.}
\Gls{noisin} is amenable to any \gls{RNN} and any noise
distribution. As with all regularization techniques, \Gls{noisin}
comes with a free parameter that determines the amount of
regularization: the spread $\gamma$ of the noise.

Certain noise distributions have bounded variance; for example the Bernoulli and the 
Beta distributions.~This limits the amount of regularization one can afford.~To circumvent 
this bounded variance issue, we rescale the noise to have unbounded variance. 
Table\nobreakspace \ref {tab:scaled_noise} shows the expression of the variance of the original noise and 
its scaled version for several distributions.  
It is the scaled noise that is used in \Gls{noisin}.  \section{Unbiased Regularization for \gls{RNN}s}\label{sec:reg}

In Section\nobreakspace \ref {sec:nirnns}, we introduced the concept of unbiasedness in the context 
of \gls{RNN}s as a desideratum for noise injection to preserve the underlying \gls{RNN}.~In this 
section we prove unbiasedness leads to an explicit regularizer 
that forces the hidden states to be robust to noise. 

\subsection{Unbiased noise injection is explicit regularization}
A valid regularizer is one that adds a nonnegative term to the risk.~This section shows 
that unbiased noise injection with exponential family likelihoods leads to valid regularizers. 

Consider the loss in Eq.\nobreakspace \ref {eq:final_objective} for an exponential family likelihood. 
The exponential family provides a general notation for the types of data encountered in 
practice: binary, count, real-valued, and categorical. Table\nobreakspace \ref {tab:A} shows the expression of $A$ for 
these types of data. The log normalizer $A(V^\top \mbz_t)$ has many useful properties. 
For example it is convex and infinitely differentiable.

Assume without loss of generality that we observe one sequence $\mbx_{1:T}$. 
Consider the empirical risk function for the noise-injected \gls{RNN}. It is defined as 
\begin{align*}
	\mathcal{R} 
	&= -\sum_{t=1}^{T}E_{p(\mbepsilon_{1:t})} \left\{ 
	 (V^\top \mbz_t)^\top \mbx_t - A(V^\top \mbz_t) \right \} + c
	.
\end{align*}
With little algebra we can decompose this risk into the sum of two terms
\begin{align}\label{eq:risk_general}
	\mathcal{R} &= \mathcal{R}\text{(det)} + 
		\sum_{t=1}^{T}E_{p(\mbepsilon_{1:t})} \left\{ \mathcal{E}_t \right \} 
\end{align}
where $\mathcal{R}\text{(det)}$ is the empirical risk for the underlying \gls{RNN} and 
$\mathcal{E}_t$ is 
\begin{align*}
	\mathcal{E}_t &= 
			A(V^\top \mbz_t) - A(V^\top \mbh_t) - \left(V^\top\mbz_t - V^\top\mbh_t \right)^\top \mbx_t
	.
\end{align*}
Because the second term in Eq.\nobreakspace \ref {eq:risk_general} is not always guaranteed to 
be non-negative, noise-injection is not explicit regularization in general. 
However, under the strong unbiasedness condition, this term corresponds to a valid 
regularization term and simplifies to
\begin{align*}
	\mathcal{R}\text{eg} &=
 	\frac{1}{2} \sum_{t=1}^{T} \mathrm{tr}\left\{
		E_{p(\mbepsilon_{1:t})} \text{Cov}(B^\top\mbz_t \g \mbz_{t-1}(\mbepsilon_{1:t-1}))
	\right \}
	,
\end{align*}
where the matrix $B = V \sqrt{\nabla^2A(V^\top\mbh_t)}$
is the prediction matrix of the underlying \gls{RNN} rescaled by the square root 
of $\nabla^2A(V^\top\mbh_t)$---the Hessian of the log-normalizer of the likelihood. 
This Hessian is also the Fisher information matrix of the \gls{RNN}. We provide a detailed 
proof in Section\nobreakspace \ref {sec:main_appendix}.

\Gls{noisin} requires that we minimize the objective of the underlying \gls{RNN} 
while also minimizing $\mathcal{R}\text{eg}$. Minimizing $\mathcal{R}\text{eg}$ induces robustness---it 
is equivalent to penalizing hidden units that are too sensitive to noise. 

\subsection{Connections}
In this section, we intuit that \Gls{noisin} has ties to ensemble methods 
and empirical Bayes.

\parhead{The ensemble method perspective.} \Gls{noisin} can be interpreted 
as an ensemble method. The objective in Eq.\nobreakspace \ref {eq:objective} corresponds to
averaging the predictions of infinitely many \gls{RNN}s at each time
step in the sequence. This is known as an ensemble method and has a
regularization effect~\citep{poggio2002bagging}. However ensemble methods are costly as they
require training all the sub-models in the ensemble. With \Gls{noisin}, at each time 
step in the sequence, one of the infinitely many \acrshort{RNN}s is trained and because 
of parameter sharing, the \acrshort{RNN} being trained at the next time step will use better
settings of the weights. This makes training the whole model efficient. (See Algorithm $1$.)

\parhead{The empirical Bayes perspective.} Consider a noise-injected
\gls{RNN}.  We write its joint distribution as
\begin{align*}
  p(\mbx_{1:T}, \mbz_{1:T}) 
  &= \prod_{t=1}^{T} p(\mbx_t | \mbz_t; V) p(\mbz_t | \mbz_{t-1}, \mbx_{t-1}; W)
\end{align*}
Here $p(\mbx_t | \mbz_t; V)$ denotes the likelihood and
$p(\mbz_t | \mbz_{t-1}, \mbx_{t-1}; W)$ is the prior over the noisy
hidden states; it is parameterized by the weights $W$. From the perspective of Bayesian inference 
this is an unknown prior. When we optimize the objective in Eq.\nobreakspace \ref {eq:objective}, we are learning 
the weights $W$. This is equivalent to learning the prior over the noisy hidden states and is known 
as empirical Bayes \citep{robbins1964empirical}. It consists in getting point estimates of prior 
parameters in a hierarchical model and using those point estimates to define the prior. \section{Empirical Study}\label{sec:empirical}

\begin{table*}[!ht]
\centering
 \captionof{table}{\Gls{noisin} improves the performance of the \gls{LSTM} and 
the dropout-\gls{LSTM} by as much as $12\%$ on the Penn Treebank dataset. This table shows word-level 
perplexity scores on the medium and large settings for both the validation (or dev) and the test set.
}\resizebox{1.0\linewidth}{!}{
\begin{tabular}{lccccccc}
\toprule
 &    \multicolumn{3}{c}{Medium} & \multicolumn{3}{c}{Large}\\\midrule
 Method & $\gamma$ & Dev & Test & $\gamma$ & Dev &  Test \\
 \midrule\midrule
None &  $--$ &  $115$ & $109$ & $--$ & $123$ & $123$ \\
Gaussian  & $1.10$ & $76.2$ & $71.8$ & $1.37$ & $73.2$ & $69.1$ \\
Logistic  & $1.06$ & $76.4$ & $72.3$ & $1.39$ & $73.6$ & $69.3$ \\
Laplace &  $1.06$ & $76.6$ & $72.4$ & $1.39$ & $73.7$ & $69.4$ \\
Gamma & $1.06$ & $78.2$ & $74.5$ & $1.39$ & $73.6$ & $69.5$ \\
Bernoulli  & $0.41$ & $\textbf{75.7}$ & $\textbf{71.4}$ & $0.33$ & $\textbf{72.8}$ & $\textbf{68.3}$\\
Gumbel  &  $1.06$ & $76.2$ & $72.7$& $1.39$ & $73.5$ & $69.5$   \\
Beta & $1.07$ & $76.0$ & $71.4$ & $1.50$ & $74.4$ & $70.2$ \\
Chi  & $1.50$ & $84.5$ & $80.7$ & $1.20$ & $79.2$ & $75.5$ \\\bottomrule
\bottomrule
\end{tabular} 
\begin{tabular}{lccccccc}
\toprule
 &    \multicolumn{3}{c}{Medium} & \multicolumn{3}{c}{Large}\\\midrule
 Method & $\gamma$ & Dev & Test & $\gamma$ & Dev &  Test \\
 \midrule\midrule
Dropout (D)  & $--$ & $80.2$ & $77.0$ & $--$ & $78.6$ & $75.3$ \\
D + Gaussian  & $0.53$ & $73.4$ & $70.4$ & $0.92$ & $\textbf{70.0}$ & $\textbf{66.1}$ \\
D + Logistic   &$0.53$ & $73.0$ & $69.9$ & $0.84$ & $69.8$ & $66.4$  \\
D + Laplace  & $0.53$ & $73.1$ & $70.0$ & $0.92$ & $69.9$ & $66.6$  \\
D + Gamma & $0.38$ & $73.5$& $70.3$ & $0.92$ & $71.1$ & $68.2$ \\
D + Bernoulli  & $0.80$ & $73.3$ & $70.1$ & $0.50$ & $\textbf{70.0}$ & $\textbf{66.1}$  \\
D + Gumbel  &  $0.46$ & $74.5$ & $71.2$ & $0.92$ & $70.2$ & $67.1$  \\
D + Beta  & $0.20$ & $\textbf{73.0}$ & $\textbf{69.2}$& $0.70$ & $70.0$ & $66.2$ \\
D + Chi & $0.29$& $76.1$ & $72.8$ & $0.82$ & $73.0$ & $70.0$  \\\bottomrule
\bottomrule
\end{tabular}} \label{tab:pp_ptb}
\end{table*}

\begin{table*}[!hbpt]
\centering
 \captionof{table}{\Gls{noisin} improves the performance of the \gls{LSTM} and 
the dropout-\gls{LSTM} by as much as $9\%$ on the Wikitext-2 dataset. This table shows word-level 
perplexity scores on the medium and large settings for both the validation (or dev) and the test set. 
D is short for dropout. $D+ \text{Distribution}$ refers to \Gls{noisin} applied to the 
dropout-\gls{LSTM} with the specified distribution.
}
\begin{tabular}{lccccccc}
\toprule
 &    \multicolumn{3}{c}{Medium} & \multicolumn{3}{c}{Large}\\\midrule
 Method & $\gamma$ & Dev & Test & $\gamma$ & Dev &  Test \\
 \midrule\midrule
None &  $--$ & $141$  & $136$ & $--$ & $176$ & $140$ \\
Gaussian  & $1.00$ & $92.7$ & $87.8$ & $1.37$ & $87.7$ & $83.4$ \\
Logistic  & $1.00$ & $93.2$ & $88.4$ & $1.28$ & $88.1$ & $83.5$\\
Laplace & $1.00$ & $95.3$ & $89.8$ & $1.28$ & $88.0$ & $83.4$ \\
Gamma &  $0.72$ & $97.6$ & $92.9$ & $1.39$ & $89.2$ & $84.5$\\
Bernoulli  & $0.54$ & $91.2$ & $86.6$ & $0.41$ & $86.9$  & $83.0$ \\
Gumbel  & $1.00$ & $95.4$ & $90.9$ & $1.28$ & $88.7$ & $84.0$  \\
Beta &  $0.80$ & $\textbf{91.1}$ & $\textbf{87.2}$ & $1.50$ & $\textbf{86.9}$ & $\textbf{82.9}$\\
Chi  &$0.20$ & $111$ & $105$ & $1.50$ & $99.0$ & $92.9$\\\bottomrule
\bottomrule
\end{tabular} 
\begin{tabular}{lccccccc}
\toprule
 &    \multicolumn{3}{c}{Medium} & \multicolumn{3}{c}{Large}\\\midrule
 Method & $\gamma$ & Dev & Test & $\gamma$ & Dev &  Test \\
 \midrule\midrule
Dropout (D)  & $--$ & $88.7$ & $84.8$ & $--$ & $95.0$ & $91.0$ \\
D + Gaussian  & $0.50$ & $86.3$ & $82.3$ & $0.69$ & $81.4$ & $77.7$ \\
D + Logistic & $0.40$ & $86.4$ & $82.5$ & $0.77$ & $81.6$ & $78.1$ \\
D + Laplace & $0.40$ & $\textbf{85.6}$ & $\textbf{82.1}$ & $0.61$ & $83.2$ & $79.1$ \\
D + Gamma & $0.30$ & $86.5$ & $82.4$& $0.61$ & $85.5$ & $81.3$ \\
D + Bernoulli  &  $0.50$ & $100.6$ & $94.4$ & $0.64$ & $\textbf{80.8}$ & $\textbf{76.8}$\\
D + Gumbel  &  $0.30$ & $86.4$ & $82.4$ & $0.53$ & $83.7$ & $80.1$ \\
D + Beta  & $0.10$ & $86.2$ & $82.3$ & $0.60$ & $81.5$ & $77.9$ \\
D + Chi  & $0.20$ & $92.0$ & $87.4$& $0.29$ & $87.1$ & $82.8$   \\\bottomrule
\bottomrule
\end{tabular} \label{tab:pp_wiki}
\end{table*}
\begin{table*}[!hbpt]
\centering
\captionof{table}{When applied to the model in
  \cite{yang2017breaking}, \Gls{noisin} achieves the same
  state-of-the-art perplexity on the Penn Treebank after only $400$ epochs (vs $1000$ 
  epochs).
  $^*$Multiplicative gamma-distributed noise with shape $2$ and scale $0.4$.
}
\begin{tabular}{lcccc}
\toprule
 Model & \# Parameters & Dev & Test \\
 \midrule\midrule
\cite{zaremba2014recurrent} - \gls{LSTM} & $20$ M & $86.2$ & $82.7$\\
\cite{gal2016theoretically} - Variational \gls{LSTM} (MC) & $20$ M & $-$ & $78.6$\\
\cite{merity2016pointer} - Pointer Sentinel-\gls{LSTM} & $21$ M & $72.4$ & $70.9$\\
\cite{grave2016improving} -  \gls{LSTM} + continuous cache pointer & $-$ & $-$ & $72.1$ \\
\cite{inan2016tying} - Tied Variational \gls{LSTM} + augmented loss & $24$ M & $75.7$ & $73.2$\\
\cite{zilly2016recurrent}- Variational RHN & $23$ M & $67.9$ & $65.4$\\
\cite{melis2017state} - 2-layer skip connection \gls{LSTM} & $24$ M & $60.9$ & $58.3$\\\bottomrule
\cite{merity2017regularizing} - AWD-\gls{LSTM} + continuous cache pointer & $24$ M & $53.9$ & $52.8$\\
\cite{krause2017dynamic} - AWD-\gls{LSTM} + dynamic evaluation & $24$ M & $51.6$ & $51.1$\\
\cite{yang2017breaking} - AWD-\gls{LSTM}-MoS + dynamic evaluation & $22$ M & $\textbf{48.3}$ & $\textbf{47.7}$\\\bottomrule
(This paper) - AWD-\gls{LSTM}-MoS + \Gls{noisin}$^*$ + dynamic evaluation & $22$ M & $\textbf{48.4}$ & $\textbf{47.6}$
 \\\bottomrule
\bottomrule
\end{tabular}\label{tab:pp_ptb_sota}
\end{table*}
We presented \Gls{noisin}, a method that relies on unbiased noise injection 
to regularize any \gls{RNN}. \Gls{noisin} is simple and can be integrated with 
any existing \gls{RNN}-based model. In this section, we focus on applying \Gls{noisin} 
to the \gls{LSTM} and the dropout-\gls{LSTM}.~We use language modeling as a testbed. 
Regularization is crucial in language modeling because the input and prediction 
matrices scale linearly with the size of the 
vocabulary. This results in networks with very high capacity. 

We used \Gls{noisin} under two noise regimes: additive noise and multiplicative noise. 
We found that additive noise uniformly performs worse than multiplicative noise 
for the \acrshort{LSTM}. We therefore report results only on multiplicative noise. 

We used \Gls{noisin} with several noise distributions: Gaussian, Logistic, Laplace, 
Gamma, Bernoulli, Gumbel, Beta, and $\chi$-Square. We found that overall the only property 
that matters with these distributions is the variance. The variance determines the amount of 
regularization for \Gls{noisin}. It is the parameter 
$\gamma$ in Algorithm $1$. We outlined in Section\nobreakspace \ref {sec:reg} how to set the noise level for a given distribution so 
as to benefit from unbounded variance.

We also found that these distributions, when used with \Gls{noisin} on the \gls{LSTM} perform 
better than the dropout \gls{LSTM} on the Penn Treebank. 

Another interesting finding is that \Gls{noisin} when applied to the dropout-\gls{LSTM} performs 
better than the original dropout-\gls{LSTM}.

Next we describe the two benchmark datasets used: Penn Treebank and Wikitext-2.~We then 
provide details on the experimental settings for reproducibility. We finally present the results 
in Table\nobreakspace \ref {tab:pp_ptb} and Table\nobreakspace \ref {tab:pp_wiki}. 

\parhead{Penn Treebank.} The Penn Treebank portion of the Wall 
Street Journal~\citep{marcus1993building} is a long standing 
benchmark dataset for language modeling. We use the standard split, where 
sections $0$  to $20$ ($930K$ tokens) are used for training, sections $21$ to $22$ 
($74K$ tokens) for validation, and sections $23$ to $24$ ($82K$ tokens) for 
testing~\citep{mikolov2010recurrent}. We use a vocabulary of size $10K$ that includes the 
special token \emph{unk} for rare words and the end of sentence indicator \emph{eos}. 

\parhead{Wikitext-2.} The Wikitext-2 dataset~\citep{merity2016pointer} has been 
recently introduced as an alternative to the Penn Treebank dataset.~It is sourced from Wikipedia 
articles and is approximately twice the size of the Penn Treebank dataset.~We use a 
vocabulary size of $30K$ and no further preprocessing steps.

\parhead{Experimental settings.} 
To assess the capabilities of \Gls{noisin} as a regularizer on its own, we used the basic 
settings for \gls{RNN} training \citep{zaremba2014recurrent}. We did not use 
weight decay or pointers \citep{merity2016pointer}. 

We considered two settings in our experiments: a medium-sized network and a large network. 
The medium-sized network has $2$ layers with $650$ hidden units each. This results 
in a model complexity of $13$ million parameters. The large network has $2$ layers 
with $1500$ hidden units each. This leads to a model complexity of $51$ million parameters.

For each setting, we set the dimension of the word embeddings to match 
the number of hidden units in each layer. Following initialization guidelines in the literature, we 
initialize all embedding weights uniformly in the interval $[-0.1, 0.1]$. All other weights 
were initialized uniformly between $[-\frac{1}{\sqrt{H}}, \frac{1}{\sqrt{H}}]$ where $H$ is 
the number of hidden units in a layer. All the biases were initialized to $0$. We fixed the seed 
to $1111$ for reproducibility. 

We train the models using truncated backpropagation through time with average stochastic 
gradient descent \citep{polyak1992acceleration} for a maximum of $200$ epochs. The \gls{LSTM} was 
unrolled for $35$ steps. We used a batch size of $80$ for both datasets. To avoid the problem 
of exploding gradients we clip the gradients to a maximum norm of $0.25$. We used an 
initial learning rate of $30$ for all experiments. This is divided by a factor of $1.2$ if the perplexity 
on the validation set deteriorates. 

For the dropout-\gls{LSTM}, the values used for dropout on the input, recurrent, and 
output layers were $0.5, 0.4, 0.5$ respectively. 

The models were implemented in PyTorch. The source code is available upon request.

\parhead{Results on the Penn Treebank.}  
The results on the Penn Treebank are illustrated in Table\nobreakspace \ref {tab:pp_ptb}. 
The best results for the non-regularized \acrshort{LSTM} correspond to 
a small network. This is because larger networks overfit and require regularization. 
In general \Gls{noisin} improves any given \acrshort{RNN} including 
dropout-\acrshort{LSTM}. For example \Gls{noisin} with multiplicative Bernoulli 
noise performs better than dropout \acrshort{RNN} for both medium and large settings. 
\Gls{noisin} improves the performance of the dropout-\gls{LSTM} by as much as $12.2\%$ 
on this dataset. 

\parhead{Results on the Wikitext-2 dataset.} 
Results on the Wikitext-2 dataset are presented in Table\nobreakspace \ref {tab:pp_wiki}. We observe 
the same trend as for the Penn Treebank dataset: \Gls{noisin} improves 
the underlying \gls{LSTM} and dropout-\gls{LSTM}. For the dropout-\gls{LSTM}, 
it improves its generalization capabilities by as much as $9\%$ on this dataset.  \section{Discussion}\label{sec:discussion}

We proposed \Gls{noisin}, a simple method for
regularizing \acrshort{RNN}s. \Gls{noisin} injects noise into the hidden 
states such that the underlying \gls{RNN} is preserved. \Gls{noisin} maximizes a 
lower bound of the log marginal likelihood of the data---the expected 
log-likelihood under the injected noise. We showed that \Gls{noisin} is an explicit 
regularizer that imposes a robustness constraint on the hidden units of 
the \gls{RNN}.~On a language modeling benchmark \Gls{noisin} improves 
the generalization capabilities of both the \gls{LSTM} and the dropout-\gls{LSTM}. 
 
\section{Detailed Derivations}\label{sec:main_appendix}
We derive in full detail the risk of \Gls{noisin} and 
show that it can be written as the sum of the risk of the original \gls{RNN} and 
a regularization term. 

Assume without loss of generality that we observe one sequence $\mbx_{1:T}$. The risk of 
a noise-injected \gls{RNN} is \begin{align*}
	\mathcal{R} 
	&= -\sum_{t=1}^{T}E_{p(\mbepsilon_{1:t})}
			\log p(\mbx_t | \mbz_{t}(\mbepsilon_{1:t})) 
	.
\end{align*}
Expand this in more detail and write $\mbz_{t}$ in lieu of $\mbz_{t}(\mbepsilon_{1:t})$ to 
avoid cluttering of notation. Then
 \begin{align*}
	\mathcal{R} 
	&= - \sum_{t=1}^{T}
		\left\{\log \nu(\mbx_t) 
	- E_{p(\mbepsilon_{1:t})}
		\left[
			 \mbz_t^\top V \mbx_t - A(V^\top \mbz_t)
		\right]\right\}
	.
\end{align*}
The risk for the underlying \gls{RNN}---$\mathcal{R}\text{(det)}$---is similar when 
we replace $\mbz_t$ with $\mbh_t$,
 \begin{align*}
	\mathcal{R}\text{(det)}
	&= - \sum_{t=1}^{T}
		\left\{\log \nu(\mbx_t) 
	- \left[
		 \mbh_t^\top V \mbx_t - A(V^\top \mbh_t)
		\right]\right\}
	.
\end{align*}
Therefore we can express the risk of \Gls{noisin} as 
a function of the risk of the underlying \gls{RNN},
 \begin{align*}\label{eq:risk}
	\mathcal{R} 
	&= \mathcal{R}\text{(det)} + \sum_{t=1}^{T} E_{p(\mbepsilon_{1:t-1})} 
		\left[E_{p(\mbepsilon_t \g \mbepsilon_{1:t-1})} \left(\mathcal{E}_1\right)\right]\\
	 \mathcal{E}_1
 	&=
		A(V^\top \mbz_t) - A(V^\top \mbh_t) - \left(V^\top\mbz_t - V^\top\mbh_t \right)^\top \mbx_t.
\end{align*}
Under the strong unbiasedness condition,
 \begin{align*}
 	E_{p(\mbepsilon_t \g \mbepsilon_{1:t-1})} \left[\mathcal{E}_1\right)
 		&= E_{p(\mbepsilon_t \g \mbepsilon_{1:t-1})} \left[A(V^\top \mbz_t) - A(V^\top \mbh_t) \right].
\end{align*}
Using the convexity property of the log-normalizer of exponential families and Jensen's inequality,
 \begin{align*}
 	E_{p(\mbepsilon_t \g \mbepsilon_{1:t-1})} \left(\mathcal{E}_1\right)
 		&\geq A(V^\top E_{p(\mbepsilon_t \g \mbepsilon_{1:t-1})}(\mbz_t)) - A(V^\top \mbh_t).
\end{align*}
Using the strong unbiasedness condition a second time we conclude
$E_{p(\mbepsilon_t \g \mbepsilon_{1:t-1})} \left(\mathcal{E}_1\right) \geq 0.$
Therefore
 \begin{align*}
 	\mathcal{R}\text{eg} &= \sum_{t=1}^{T} E_{p(\mbepsilon_{1:t-1})} 
		\left[E_{p(\mbepsilon_t \g \mbepsilon_{1:t-1})} \left(\mathcal{E}_1\right)\right] \geq 0
\end{align*}
is a valid regularizer.~A second-order Taylor expansion of $A(V^\top \mbz_t)$ around $A(V^\top \mbh_t)$ 
and the strong unbiasedness condition yield 
 \begin{align*}
 	\mathcal{R}\text{eg} &= \frac{1}{2}\sum_{t=1}^{T} \mathrm{tr}\left\{
		E_{p(\mbepsilon_{1:t-1})} 
			\left[
			\text{Cov}(B^\top\mbz_t \g \mbz_{t-1}(\mbepsilon_{1:t-1}))
			\right]
		\right\},
\end{align*}
where the matrix $B = V\sqrt{\nabla^2A(V^\top\mbh_t)}$ is the original 
prediction matrix $V$ rescaled by the square root of the Hessian of 
the log-normalizer, the inverse Fisher information matrix of the underlying \gls{RNN}.
This regularization term forces the hidden units to be robust to noise. 
Under weak unbiasedness, the proof holds under the assumption that 
the true data generating distribution is an \gls{RNN}.  

\section*{Acknowledgements} 
We thank Francisco Ruiz for presenting our paper at ICML, 2018.
We thank the Princeton Institute for Computational Science and Engineering (PICSciE), the 
Office of Information Technology's High Performance Computing Center 
and Visualization Laboratory at Princeton University for the computational resources.
This work was supported by ONR N00014-15-1-2209, ONR 133691-5102004, 
NIH 5100481-5500001084, NSF CCF-1740833, the Alfred P. Sloan Foundation, 
the John Simon Guggenheim Foundation, Facebook, Amazon, and IBM.

\bibliography{noisin}
\bibliographystyle{icml2018}

\end{document}